\documentclass[a4paper,twoside]{article}

\usepackage{epsfig}
\usepackage{subcaption}
\usepackage{calc}
\usepackage{amssymb}
\usepackage{amstext}
\usepackage{amsmath}
\usepackage{amsthm}
\usepackage{multicol}
\usepackage{pslatex}
\usepackage{apalike}
\usepackage{algorithm2e}
\usepackage[bottom]{footmisc}
\usepackage{SCITEPRESS}

\begin{document}

\title{Machine Learning in Industrial Quality Control of Glass Bottle Prints}

\author{\authorname{Maximilian Bundscherer\sup{1}, Thomas H. Schmitt\sup{1} and Tobias Bocklet\sup{1}}
\affiliation{\sup{1}Department of Computer Science, Technische Hochschule Nürnberg Georg Simon Ohm, Nuremberg, Germany}
\email{\{maximilian.bundscherer, thomas.schmitt, tobias.bocklet\}@th-nuernberg.de}
}

\keywords{Machine Learning, Quality Control, Industrial Manufacturing, Glass Bottle Prints}

\abstract{
In industrial manufacturing of glass bottles, quality control of bottle prints is necessary as numerous factors can negatively affect the printing process.
Even minor defects in the bottle prints must be detected despite reflections in the glass or manufacturing-related deviations.
In cooperation with our medium-sized industrial partner, two ML-based approaches for quality control of these bottle prints were developed and evaluated, which can also be used in this challenging scenario.
Our first approach utilized different filters to supress reflections (e.g. Sobel or Canny) and image quality metrics for image comparison (e.g. MSE or SSIM) as features for different supervised classification models (e.g. SVM or k-Neighbors), which resulted in an accuracy of \(84\%\).
The images were aligned based on the ORB algorithm, which allowed us to estimate the rotations of the prints, which may serve as an indicator for anomalies in the manufacturing process.
In our second approach, we fine-tuned different pre-trained CNN models (e.g. ResNet or VGG) for binary classification, which resulted in an accuracy of \(87\%\).
Utilizing Grad-Cam on our fine-tuned ResNet-34, we were able to localize and visualize frequently defective bottle print regions.
This method allowed us to provide insights that could be used to optimize the actual manufacturing process.
This paper also describes our general approach and the challenges we encountered in practice with data collection during ongoing production, unsupervised preselection, and labeling.
}

\onecolumn \maketitle \normalsize \setcounter{footnote}{0} \vfill

\section{\uppercase{Introduction}}
\label{lbl_Introduction}

\begin{figure}
	\centerline{
		\includegraphics[width=0.321\linewidth]{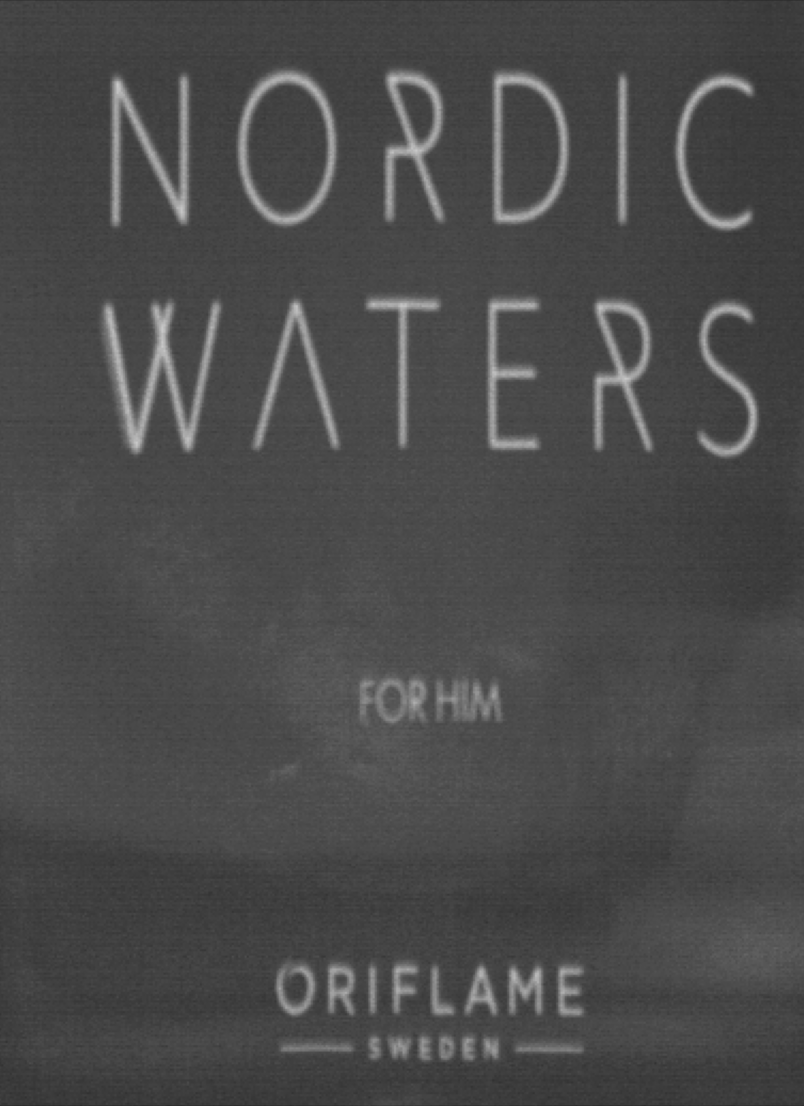}\hfill
		\includegraphics[width=0.321\linewidth]{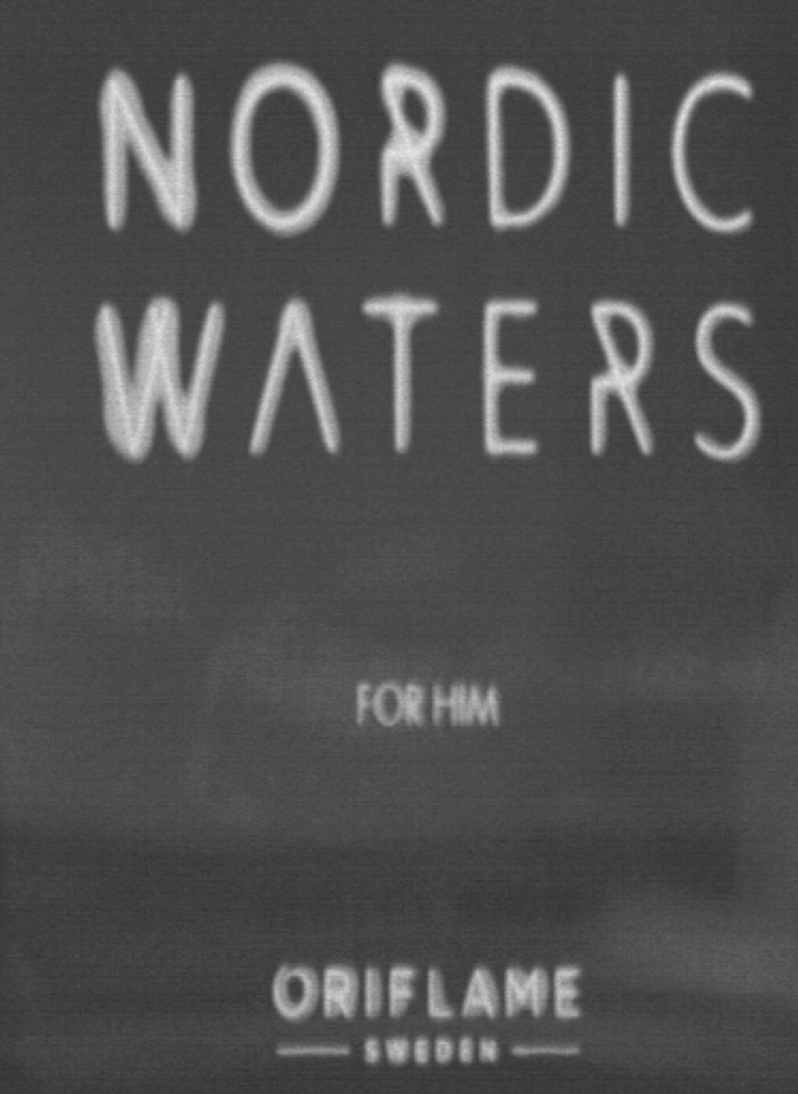}\hfill
		\includegraphics[width=0.321\linewidth]{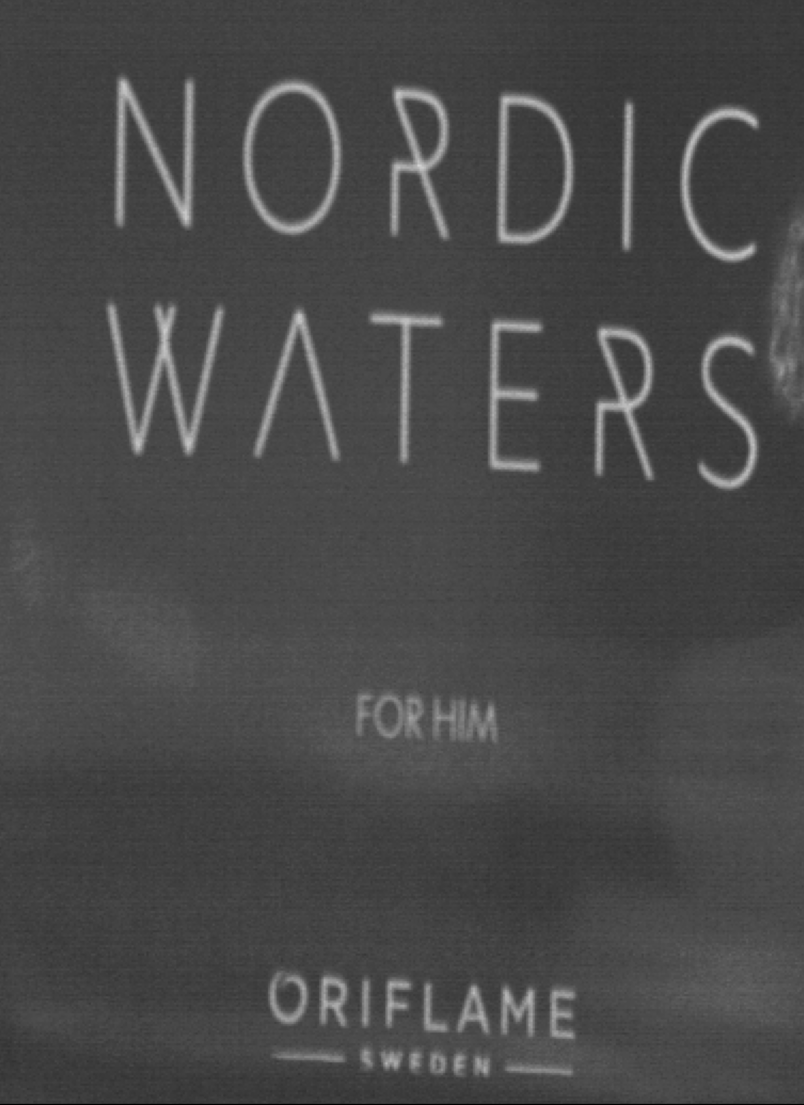}
	}
	\caption{
		Cropped images of glass bottle prints: (left) acceptable print, (middle) unacceptable smeared print, and (right) unacceptable rotated print.
	}
	\label{fig_Defects}
\end{figure}

In industrial manufacturing, glass can be printed utilizing a technique called Silk-Screen Printing.
In the Silk-Screen Printing process, the ink is transferred onto the glass surface through a stencil.
The stencils must be constantly adjusted during ongoing production, as the stencils wear out in various areas and the quality of the prints is negatively affected as a result.
For instance, the prints could be smeared, incomplete, or rotated, as shown in Figure \ref{fig_Defects}.
In cooperation with our medium-sized industrial partner, who also relies on Silk-Screen Printing, we investigated the quality control of glass bottle prints utilizing machine learning and computer vision methods.

\subsection{Our Contributions}
\label{lbl_Introduction_Our_Contributions}

The main contributions of this work are:
\begin{itemize}
	\item Data collection during ongoing production.
	\item Unsupervised preselection of images and image labeling by quality assurance (QA) experts.
	\item Development and evaluation of an approach (AP1) based on a reference image of an acceptable print, ORB alignment, filters, image quality metrics, and supervised classification models.
	\item Development and evaluation of a second approach (AP2) based on pre-trained CNN models.
	\item Proposal of a method in which ORB image alignment parameters, such as rotation, can be used as an anomaly indicator.
	\item Proposal of a method in which Grad-Cam visualizations of our fine-tuned CNN models can be employed to identify frequently defective areas of bottle prints.
\end{itemize}

\subsection{Related Work}
\label{lbl_Introduction_Releated_Work}

In \cite{zhou2019automated}, defects in glass bottle bottoms were classified using Saliency Detection and Template Matching.
The comprehensive study \cite{zhou2019surface} investigated Visual Attention Models and Wavelet Transformations for the same purpose.
While these studies did not address the classification of glass bottle prints, they are related to the quality control of glass bottles in industrial processes.
We employed pre-trained CNN models, which have also been utilized in other industrial applications studies.
In \cite{villalba2019deep}, CNN models for quality control in the printing industry were investigated.
CNN models were also evaluated for quality control of textures in the automotive industry \cite{malaca2019online}.
At supermarket checkouts, VGG models from \cite{hossain2018automatic} were able to classify fruits automatically, and ResNet models from \cite{quach2020identification} were suitable for detecting chicken diseases.
CNN models were also investigated in pathological brain image classification in \cite{kaur2019automated}.

\subsection{Outline}
\label{lbl_Introduction_Outline}

This paper is structured as follows:
Section \ref{lbl_Data} describes, the actual application \& our data and how we collected data during ongoing production and made an unsupervised pre-selection for data labelling.
Section \ref{lbl_Feature_Extraction} describes how the images were prepared for classification, comprising filters, image quality metrics (IQMs), and image alignment based on the ORB algorithm.
In Section \ref{lbl_Approaches}, our two approaches are independently introduced.
In Section \ref{lbl_Results}, we evaluate our approaches and show how our methods can be used beyond binary classification for industrial manufacturing process optimization.
In Section \ref{lbl_Discussion}, we interpret our results and describe the limitations.
The final Section \ref{lbl_Conclusion} presents the findings of this study

\section{\uppercase{Application \& Data}}
\label{lbl_Data}
Reflections in glass bottles are unavoidable when taking pictures with an (external) light source.
Glass reflections that occur during photography are described in \cite{sudo2021detection}.
In industrial manufacturing, attempts are made to avoid this challenge by adjusting the image-capturing process so that the reflections are shifted to regions that are irrelevant for quality control.

Our partner already uses an automatic system for quality control, which does not reliably detect defective bottle prints and is not robust against environmental influences, like reflections and misalignment.
The currently used system compares bottle images with a reference image of an acceptable bottle and counts the number of differing pixels.
How far the bottle deviates from the reference bottle to still be classified as acceptable has to be specified via a threshold value.
Since reflections lead to more variation than the actual defects, the system is inadequate for bottles where it is impossible to cause the reflections to occur outside the bottle prints in the image-capturing process.

Dispatched defective bottles cost the company significantly more in take-back costs and reputational damage than rejecting acceptable bottles.
So, our partner specified that the focus should be on a high true positive rate since manual follow-up inspections of rejected bottles can mitigate false positives.
Since one classification system is used per production line, a defective bottle must be classified and sorted out within \(1s\).
The mechanism to remove defective bottles takes about \(0.2s\), which entails that our system must be able to process images within \(0.8s\).

During the initial data analysis, two additional challenges were identified:
A major challenge is that bottles do not have exactly the same shape due to micro variations in the manufacturing process.
Therefore, the reflections in the glass bottles always occur in different regions.
Minimal differences in glass shapes are otherwise negligible for the quality control of bottle prints.
Another challenge is that the bottles are always slightly displaced or rotated in the images due to the physical conditions of photography (rotating axis).
It is difficult to determine whether the bottle or the print is rotated or shifted, which would indicate a defective bottle print.
The current system does not use automatic image alignment, which, together with the various occurring reflections, explains the unreliable classification results.

\subsection{Data Collection}
\label{lbl_Data_Collection}

\begin{figure}
	\centerline{\includegraphics[width=1\linewidth]{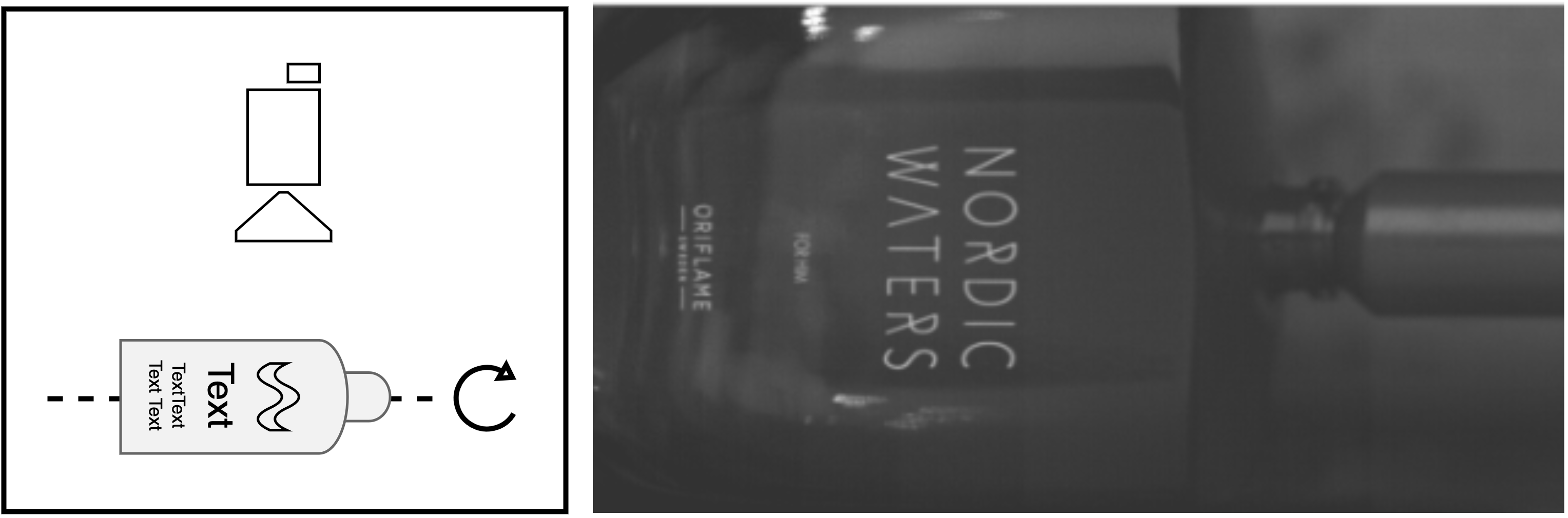}}
	\caption{	
		Photography setup: (left) system overview and (right) a captured image. The camera and lighting remain stationary during the capture, and the bottle is rotated.
	}
	\label{fig_Camera_Setup}
\end{figure}
Our industry partner already automatically captures images with a fully integrated system.
Each bottle is photographed at the end of the production line. In addition to the time stamp, it is also stored whether the bottle is classified as acceptable according to the classification system currently in use.
The bottles are held and rotated with a rotating axis for photography, keeping the camera and the lighting stationary, as shown in Figure \ref{fig_Camera_Setup}.

We copied the images during ongoing production, which was not allowed to be disturbed.
To be independent of an internet connection, we continuously copied the images and metadata to an external 4TB USB 3.0 hard drive from a Samba share via Raspberry Pi 4, rsync and cron.
Transferring all images in real time was impossible due to network connectivity and hardware limitations on read and write speeds.
After some practical tests, we decided to copy a maximum of \(500\) images (approx. \(5GB\)) every \(15m\).
The less frequently bottle images, considered unacceptable by the current system, have been prioritized.

\subsection{Unsupervised Preselection}
\label{lbl_Data_Unsupervised_Preselection}

Reducing the number of images was necessary for the labeling process. Due to time and cost constraints, a manual review of all images was not possible, so we decided on an unsupervised preselection.
In our preselection, the bottle images were compared with a reference image of an acceptable bottle. Our QA experts specified this reference bottle.
Image quality metrics (IQMs) were utilized for these comparisons, described in Section \ref{lbl_Feature_Extraction_Image_Quality_Metrics}.
The comparison window was chosen as small as possible to reduce the influence of reflections.
We managed to compare only the actual prints without reflections.
The images were aligned based on the ORB algorithm to mitigate the influence of physical conditions during photography, see Section \ref{lbl_Feature_Extraction_ORB_Alignment}.

Images that differed more than \(80\%\) from the average (potentially unacceptable prints) were approved for labeling together with images that differed less than \(20\%\) from the average (potentially acceptable prints). These values were adjusted to represent an expected manufacturing distribution based on the expertise of our industry partner.

\subsection{Labeling}
\label{lbl_Data_Labeling}

Our QA experts labeled \(800\) bottle print images using LabelStudio \cite{LabelStudio}.
A customized labeling template was built to rate rotated, smeared, cropped, shifted, or incomplete prints using a scale from \(1\) to \(6\).
These ratings were mapped to binary labels (acceptable and unacceptable prints) due to the future application and widely varying defects.
It was also challenging for our QA experts to label the defects meaningfully according to the abovementioned criteria, so they often used the other defect option.
Our QA experts specified a print as unacceptable if any defect was rated higher than \(3\).

In total, \(83\) bottles were specified as unacceptable.
Therefore \(166\) (\(+83\) images of acceptable prints) were used for the evaluation.

\section{\uppercase{Data Preparation}}
\label{lbl_Feature_Extraction}

We utilized colored images (\(3\) channels) or grayscale images (\(1\) channel) for our experiments.
We also investigated different data preparation techniques, comprising image quality metrics (IQMs), filters, and image alignment based on the ORB algorithm.

\subsection{Image Quality Metrics}
\label{lbl_Feature_Extraction_Image_Quality_Metrics}

Image quality metrics (IQMs) are measures for assessing the similarity of two images and are usually used to evaluate image compression algorithms \cite{sara2019image} \cite{bakurov2022structural} \cite{tan2013perceptually} \cite{jain2011full}.
We utilized IQMs to compare an image of a test bottle with an image of a reference bottle. A large difference should indicate that the bottle print is unacceptable.

This study utilized \textit{Mean-Squared Error (MSE)}, \textit{Normalized Root MSE (NRMSE)}, and \textit{Structural Similarity Index (SSIM)} \cite{sara2019image} as IQMs for image comparison.

\subsection{Image Filters}
\label{lbl_Feature_Extraction_Image_Filters}

\begin{figure}
	\centerline{\includegraphics[width=1\linewidth]{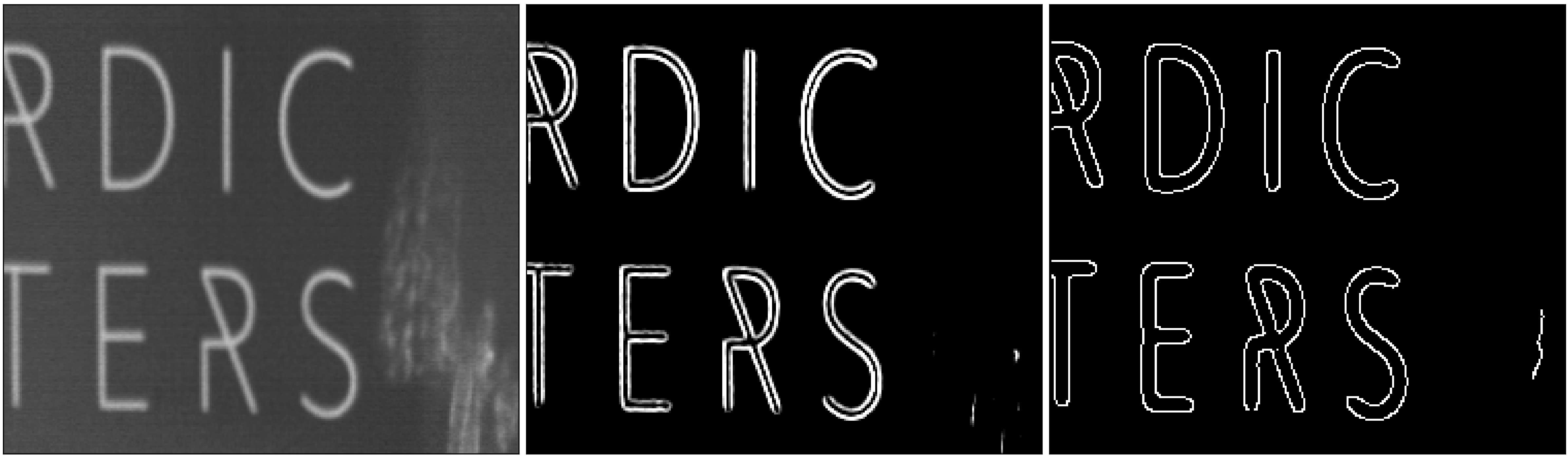}}
	\caption{
		Cropped images: (left) image without filter, (middle) image with Sobel, and (right) image with Canny. Reflections and the luminous background are reduced by applying these filters.
	}
	\label{fig_Filters}
\end{figure}
Various reflections lead to more variation than the actual defects.
Edge detectors (or filters) such as Sobel \cite{gonzales1987digital} and Canny \cite{canny1986computational} are able to reduce the effects of reflections in images by emphasizing strong edges and suppressing weak edges \cite{ozturk2015comparison} \cite{forcado2018model}.
We tested different filters and parameters to emphasize bottle print letters and suppress reflections, as shown in Figure \ref{fig_Filters}.

This study utilized \textit{Sobel}, \textit{Sobel-v}, \textit{Sobel-h}, \textit{Canny-2}, \textit{Canny-2.5}, and \textit{Canny-3} filters.
For Sobel filters, the suffixes refer to their directions (vertical and horizontal). For Canny, the suffixes refer to the sigma values used (\(2\), \(2.5\), and \(3\)).
We also utilized the original images \textit{No-Filter} and histogram equalized images \textit{Equal-Hist}.

\subsection{ORB Alignment}
\label{lbl_Feature_Extraction_ORB_Alignment}

The bottles are always slightly displaced or rotated in the images due to the physical conditions of photography.
For image comparison utilizing IQMs, we relied on aligned images based on the ORB algorithm.
The ORB (Oriented FAST and Rotated BRIEF) algorithm is based on the Oriented FAST (Features from Accelerated Segment Test) Corner Detection algorithm and the Rotated BRIEF (Binary Robust Independent Elementary Features) Descriptor \cite{rublee2011orb}.
The reference image of an acceptable bottle specified by our QA experts was used as alignment reference for all other images.

The rotation of an image related to a reference image can be estimated by the detected keypoints (BRIEF Descriptors) and computations on the homography matrix \cite{luo2019overview}.
We use this technique to get a better understanding of rotated bottle prints in industrial manufacturing over time.

\section{\uppercase{Approaches}}
\label{lbl_Approaches}

\begin{figure}
	\centerline{\includegraphics[width=1\linewidth]{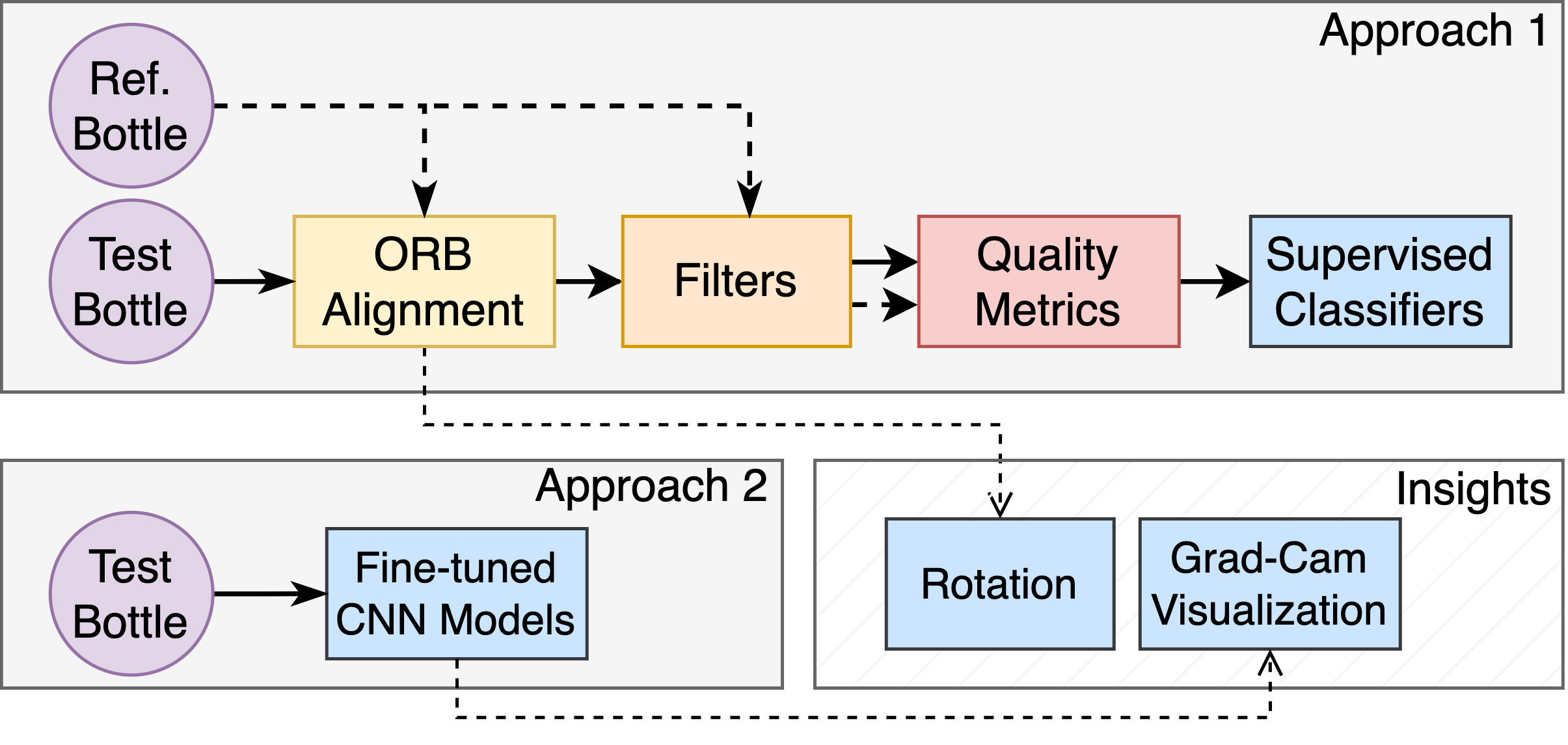}}
	\caption{
		Visualization of our two approaches: (AP1) ORB, Filters, IQMs, \& Classifiers and (AP2) Transfer Learning with CNN Models.
		Some methods from our approaches can be used beyond binary classification (Insights).
	}
	\label{fig_Abstract_Approaches}
\end{figure}
Figure \ref{fig_Abstract_Approaches} shows an abstract visualization of our two approaches AP1 and AP2.
We relied on Leave-One-Out Cross-Validation (LOOCV) to train and evaluate our models.
In total, \(83\) bottles were specified as unacceptable.
Therefore \(166\) (\(+83\) images of acceptable prints) were used utilized.

\subsection{AP1: ORB, Filters, IQMs, and Classifiers}
\label{lbl_Approaches_AP1}

In our first approach (AP1), the test bottle images were aligned based on the ORB algorithm via the reference image of an acceptable bottle, see Section \ref{lbl_Feature_Extraction_ORB_Alignment}.
In the next step, \(8\) filters were applied to the test bottle images and to the reference image (for subsequent image comparisons), see Section \ref{lbl_Feature_Extraction_Image_Filters}.
Finally, \(3\) IQMs were utilized for comparing the processed test bottle images with the processed reference images, see Section \ref{lbl_Feature_Extraction_Image_Quality_Metrics}.

In total, \(24\) combinations (\(8\) filters $\times$ \(3\) IQMs) were utilized as features for our \textit{SVM}, \textit{k-nearest Neighbors}, \textit{Random Forest}, \textit{Decision Tree} and \textit{Neuronal Network} classifiers.

\subsection{AP2: Transfer Learning with CNN Models}
\label{lbl_Approaches_AP2}

In our second approach (AP2), we fine-tuned CNN models pre-trained on ImageNet \cite{deng2009imagenet} for binary classification.
We used ResNet and VGG models, which have also been utilized in other studies to classify rail \cite{song2020resnet} and steel \cite{abu2021performance} defects and in which they also had to handle light reflections.
We additionally tested AlexNet \cite{krizhevsky2017imagenet}, and DenseNet \cite{huang2017densely}.

This approach employed standard machine learning techniques, including Binary Cross Entropy (BCE) as loss function, Sigmoid as activation function, Early Stopping, StepLR as learning rate scheduler, and Adam as optimizer.
For comparison, we additionally froze the layers' weights preceding our custom binary classification head during training.

This study fine-tuned \textit{ResNet-18}, \textit{ResNet-34}, \textit{ResNet-50}, \textit{ResNet-101}, \textit{ResNet-152}, \textit{VGG-11}, \textit{VGG-13}, \textit{VGG-16}, \textit{AlexNet} and \textit{DenseNet-121}.

We additionally utilized Grad-Cam \cite{selvaraju2017grad}, a visual neural network explanation method, like Score-CAM \cite{wang2020score} or LFI-CAM \cite{lee2021lfi}, to provide heat maps highlighting regions of an image that are important for classification.

\section{\uppercase{Results and Evaluation}}
\label{lbl_Results}

\begin{figure}
	\centerline{
		\centerline{\includegraphics[width=0.9\linewidth]{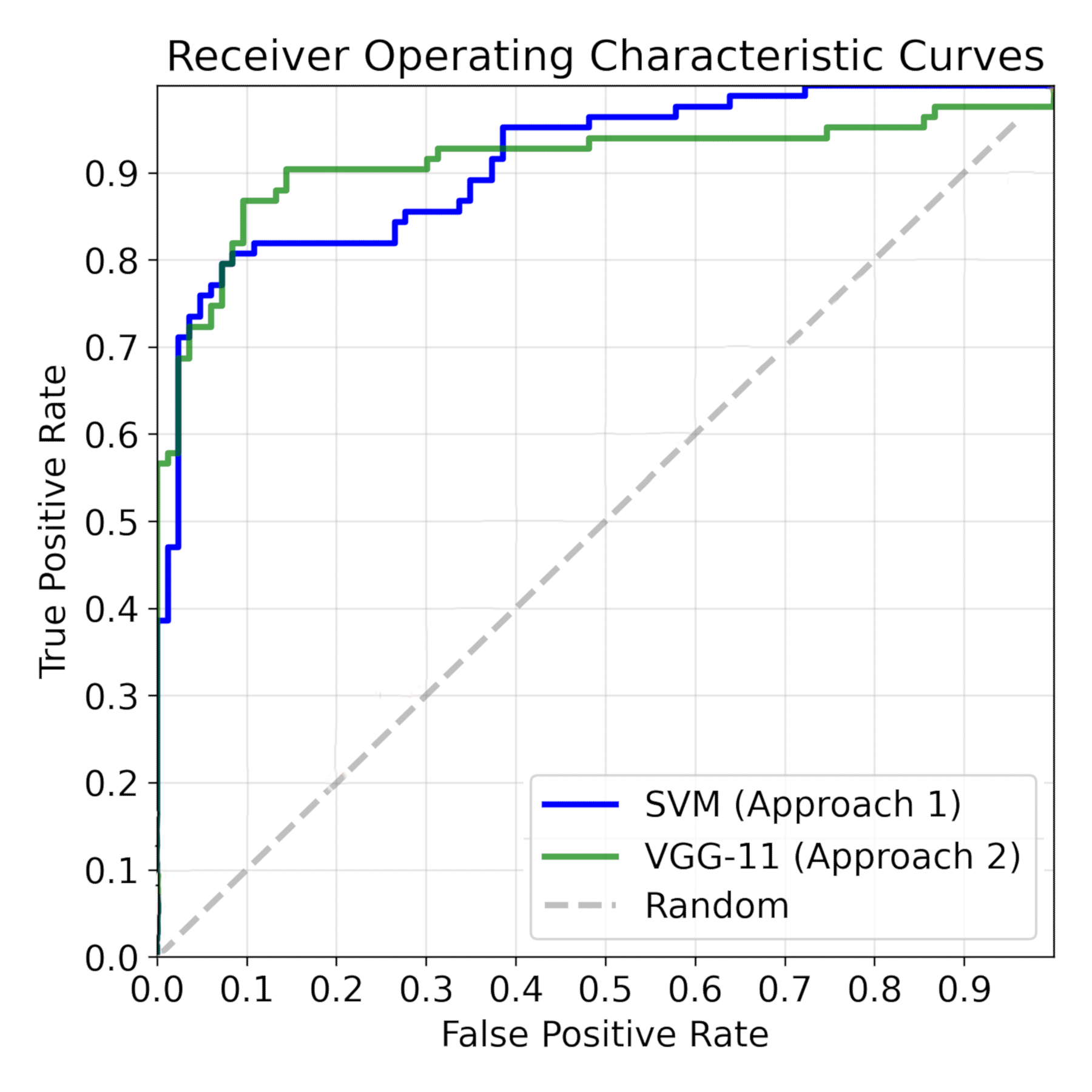}}
	}
	\caption{
		ROC Curves of our most accurate models: SVM (AP1) and VGG-11 (AP2).
	}
	\label{fig_Roc_Curves}
\end{figure}

In this study, true positives (\(TP\)) represent correctly classified unacceptable bottle prints, and true negatives (\(TN\)) represent correctly classified acceptable bottle prints. It is also necessary to consider false positives (\(FP\)), where acceptable prints are incorrectly classified as unacceptable, and false negatives (\(FN\)), where unacceptable prints are incorrectly classified as acceptable.
It is advisable to utilize the true positive rate (or sensitivity) for better interpretability and especially when the underlying class distribution is unbalanced.
The true negative rate is called specificity.
In addition to these metrics, we also includes receiver operating characteristic (ROC) Curves to visualize the trade-off between \(TP\) and \(FP\) rates across our classification thresholds, shown in Figure \ref{fig_Roc_Curves}.

\subsection{Baseline Approach}
\label{lbl_Results_Current_Approach}

To train and evaluate our models, we relied on labels provided by our QA experts.
The currently used approach classified our images with an accuracy of \(66\%\) (\(28~TP\), \(1~FP\), \(82~TN\), \(55~FN\)).
The sensitivity is \(34\%\), and the false positive rate is \(1\%\).

\subsection{AP1: ORB, Filters, IQMs, and Classifiers}
\label{lbl_Results_AP1}

\begin{table}
	\caption{
		Accuracy, \(TP\), \(FP\), \(TN\) and \(FN\) of our five supervised classifiers from AP1 (Section \ref{lbl_Approaches_AP1}).
	}
	\begin{tabular}{p{28mm}p{2.5mm}p{2.5mm}p{2.5mm}p{2.5mm}p{12mm}}
		\hline
		\textbf{Classifier}             & \textbf{TP}  & \textbf{FP}  & \textbf{TN}  & \textbf{FN}  & \textbf{Accuracy} \\ \hline\hline
		\textbf{SVM} 					& 61 & 4 & 79 & 22 & 84.34\%    \\
		\textbf{k-NN}             & 68 & 12 & 71 & 15 & 83.73\%    \\
		\textbf{Random Forest}                    & 67 & 13 & 70 & 16 & 82.53\%    \\
		\textbf{Decision Tree}          & 71 & 18 & 65 & 12 & 81.93\%    \\
		\textbf{Neuronal Network}          & 75 & 38 & 45 & 8 & 72.29\%    \\ \hline
	\end{tabular}
	\label{tab_Approach_1_Results}
\end{table}
As shown in Table \ref{tab_Approach_1_Results}, SVM achieved the highest accuracy of \(84\%\) in AP1, see Section \ref{lbl_Approaches_AP1}.
The sensitivity is \(73\%\), and the false positive rate is \(5\%\).
Figure \ref{fig_Roc_Curves} shows the ROC Curve of this model.
Without aligning the images based on the ORB algorithm, the accuracy decreased by \(18\%\) on average.
The average accuracy also decreased by \(2\%\) when colored images (\(3\) channels) were utilized for classification instead of \(1\) channel images.
Mapping \(3\) channels into \(1\) channel by a weighted sum of individual channels or utilizing only the green channel of the images had no noticeable effect on our models' accuracy.

\subsection{AP2: Transfer Learning with CNN Models}
\label{lbl_Results_AP2}

\begin{table}
	\caption{
		Accuracy, \(TP\), \(FP\), \(TN\) and \(FN\) of our ten fine-tuned CNN models from AP2 (Section \ref{lbl_Approaches_AP2}).
	}
	\begin{tabular}{p{28mm}p{2.5mm}p{2.5mm}p{2.5mm}p{2.5mm}p{12mm}}
		\hline
		\textbf{Model}     & \textbf{TP}  & \textbf{FP}  & \textbf{TN}  & \textbf{FN}  & \textbf{Accuracy} \\ \hline\hline
		\textbf{VGG-11}  & 70 & 8 & 75 & 13 & 87.35\%    \\
		\textbf{VGG-13} & 68 & 7 & 76 & 15 & 86.75\%    \\
		\textbf{VGG-16}   & 70 & 9 & 74 & 13 & 86.75\%    \\ \hline
		\textbf{ResNet-18}   & 65 & 9 & 74 & 18 & 83.73\%    \\
		\textbf{ResNet-34}     & 67 & 8 & 75 & 16 & 85.54\%    \\
		\textbf{ResNet-50}     & 57 & 6 & 77 & 26 & 80.72\%    \\
		\textbf{ResNet-101}    & 58 & 4 & 79 & 25 & 82.53\%    \\
		\textbf{ResNet-152}     & 62 & 4 & 79 & 21 & 84.94\%  \\ \hline
		\textbf{AlexNet}  & 66 & 11 & 72 & 17 & 83.13\%    \\
		\textbf{DenseNet-121}  & 63 & 8 & 75 & 20 & 83.13\%    \\ \hline
	\end{tabular}
	\label{tab_Approach_2_Results}
\end{table}
As shown in Table \ref{tab_Approach_2_Results}, VGG-11 achieved the highest accuracy of \(87\%\) in AP2, see Section \ref{lbl_Approaches_AP2}.
The sensitivity is \(84\%\), and the false positive rate is \(10\%\).
Figure \ref{fig_Roc_Curves} shows the ROC Curve of this model.
Aligning the images based on the ORB algorithm decreased accuracy by \(5\%\) on average.
We were unable to train non-pre-trained ResNet and VGG models with an average accuracy of \(52\%\).
We have too few images of defective prints to train such large models from scratch.
Training with frozen model weights decreased the average accuracy by \(17\%\).

\subsection{Insights for Manufacturing}
\label{lbl_Results_Insights_Manufacturing}

\begin{figure}[h]
	\centerline{\includegraphics[width=1\linewidth]{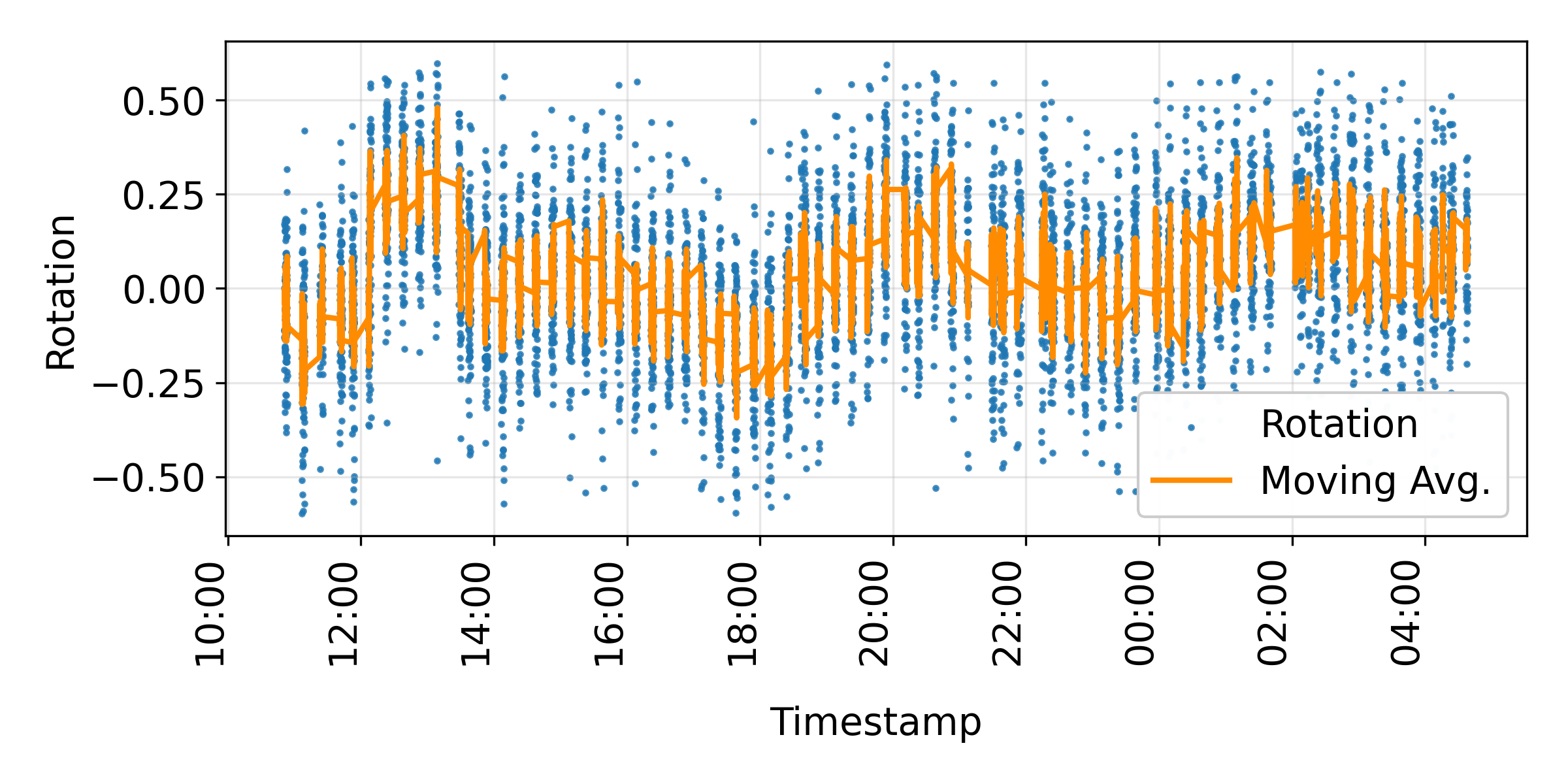}}
	\caption{
		Estimated rotations based on ORB image alignment plotted over time. At about 12 o'clock, the print stencil was replaced.
	}
	\label{fig_Rotations}
\end{figure}

Two methods we used to analyze our trained models also seem suitable for optimizing the actual manufacturing process:

The rotation of an image related to a reference image can be estimated by image alignment based on the ORB algorithm.
Over time, the rotations follow a sinusoidal pattern, as shown in Figure \ref{fig_Rotations}.
According to our QA experts, this pattern can be explained by manufacturing-related deviations.
Ongoing anomaly detection might be possible by monitoring deviations from this expected pattern.

\begin{figure}[h]
	\centerline{
		\includegraphics[width=0.485\linewidth]{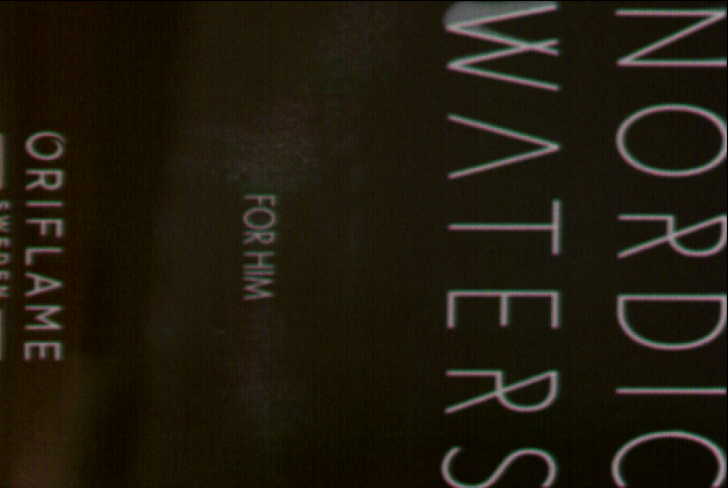}\hfill
		\includegraphics[width=0.485\linewidth]{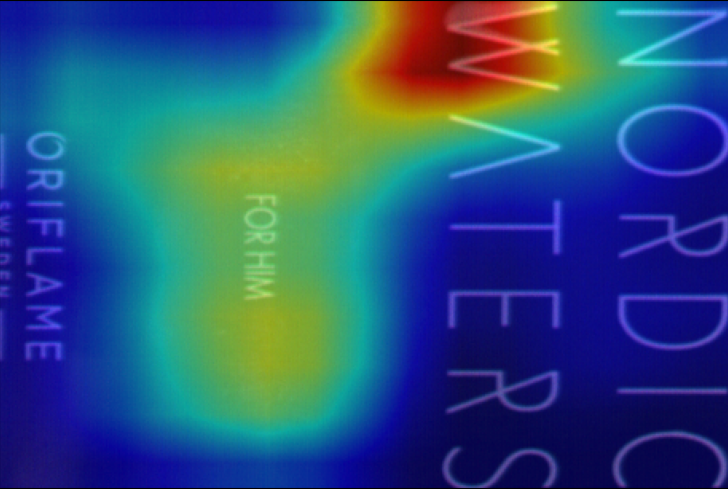}
	}
	\caption{
		Grad-Cam visualization of our fine-tuned ResNet-34 model: (left) original image of an unacceptable print and (right) Grad-Cam heat map.
		These heat maps highlight regions of an image that are important for classification.
		The CNN model focused on this example's smeared letter \textit{W}.
	}
	\label{fig_GradCam}
\end{figure}

We also utilized Grad-Cam to provide heat maps highlighting regions of an image important for classification for our fine-tuned CNN models.
We observed that heat maps from our fine-tuned ResNet-34 model are correlated with the actual defects in an image.
An example is shown in Figure \ref{fig_GradCam}.
By averaging the heat maps across all images of unacceptable prints, we were able to localize and visualize frequently defective bottle print regions.
Therefore, it was possible to identify spots where the print stencil in the production line should be enhanced.

\section{\uppercase{Discussion}}
\label{lbl_Discussion}

In this Section, the results of our two approaches are discussed separately.
Following this, our approaches are compared with the currently used classification system and the requirements of the industrial application.
In the last Subsection, the limitations of our methods are addressed.

\subsection{AP1: ORB, Filters, IQMs, and Classifiers}
\label{lbl_Discussion_AP1}

AP1 achieved an accuracy of \(84\%\).
This approach relies on aligned images for meaningful classifications.
We observed that our classifiers are more accurate on \(1\) channel images than on \(3\) channels images.
Utilizing only green channels (\(1\) channel images) of the bottle print images is proposed to reduce computation time and increase accuracy.

Although this approach is based on supervised classifiers, it could also be interpreted as an anomaly detection method due to our data preparation strategy:
These supervised classifiers are not trained to classify images of unacceptable and acceptable prints directly.
Instead, they are trained on how much a bottle print may differ from an acceptable reference bottle print to be still classified as acceptable.

\subsection{AP2: Transfer Learning with CNN Models}
\label{lbl_Discussion_AP2}

AP2 achieved the highest accuracy of \(87\%\).
In contrast to AP1, the fine-tuned CNN models are able to classify our images of the prints directly.
This approach can be interpreted as replacing the fixed filters with CNN-Convolutions.
As a result, AP2 relies on fewer potentially false assumptions than AP1.

Image alignment based on the ORB algorithm decreased the average accuracy of these CNN models.
The tested CNN models were pre-trained on colored images (\(3\) channels) without any applied filters.
We achieved the best accuracy with unaligned original images of bottle prints and unfrozen model weights.

\subsection{Industrial Application}
\label{lbl_Discussion_Industrial_Application}

The currently deployed approach classified our images with an accuracy of \(66\%\).
Compared to these results, our two approaches correctly classified more unacceptable prints (\(+ 51\%\)), with a higher false positive rate (\(+ 8\%\)) (AP2).
This comparison is not particularly meaningful because the currently used approach is primarily influenced by various reflections and misalignment, which is why a smaller comparison window was chosen for the current approach.
This was necessary in order to be able to use the currently employed approach for classification in practice.

Our QA experts considered the sensitivities and false positive rates of our two approaches as acceptable for quality control.
Further steps are required to reduce the effort in the manual follow-up inspections of incorrectly rejected acceptable bottles.

Our approaches proved robust against various reflections in our experiments.
It was possible to mitigate the influence of reflections by applying filters in AP1.
Our tested image alignment method is not strongly influenced by reflections due to the ORB keypoint detection mechanism.
Our fine-tuned CNN models from AP2 proved robust against reflections without additional steps.

All our approaches met the industrial application time constraint of maximum \(0.8s\).
Image preparation and classification could be performed in about maximum \(0.4s\) per image (tested on a NVIDIA GeForce RTX 2080 Ti consumer graphic card).

\subsection{Limitations}
\label{lbl_Discussion_Limitations}

We were unable to copy every image during ongoing production.
It was also uneconomical to label every image, so we opted for an unsupervised preselection.
Therefore, our preselected images may not be fully representative.
We utilized empirical information from our industry partner, such as the expected proportion of unacceptable bottles, to improve our preselection strategy.

AP1 relies on aligned images for accurate classification.
The models may be unable to classify if the bottle print was already rotated before alignment, indicating an unacceptable print.
This challenge could be addressed by adding the alignment parameters as features for classification.
However, due to our discontinuous data collection and since only a few bottles had rotated prints, we were unable to separate process-related variations from the actual bottle print rotations.
Therefore, we were also unable to evaluate our proposed method for anomaly detection based on computed rotations.

We observed that image alignment decreased the accuracy of our tested CNN models.
Further research is necessary to determine whether this is due to general image alignment or the ORB algorithm's use.

Our dataset is highly imbalanced and contains only a few images of widely varying unacceptable prints.
Therefore, we relied on binary labels due to the future application and these widely varying defects.

\section{\uppercase{Conclusion}}
\label{lbl_Conclusion}
We recommend AP2, utilizing fine-tuned CNN models, such as ResNet or VGG.
We achieved the highest accuracy of \(87\%\) with the help of VGG-11, see Section \ref{lbl_Results_AP2}.
This approach has the highest accuracy of our tested approaches and relies on minimal assumptions, which is less likely to lead to errors.

The tested CNN models proved robust against reflections without any additional steps.
Utilizing our fine-tuned ResNet-34 model and Grad-Cam, we were able to localize and visualize spots where the print stencil in the production line should be enhanced, see Section \ref{lbl_Results_Insights_Manufacturing}.

Although we achieved a slightly lower accuracy of \(84\%\) with models from AP1, methods of this approach offer some advantages:
Selecting a specific filter or IQM can help prioritize detecting certain defects or reduce the influence of reflections, see Section \ref{lbl_Feature_Extraction_Image_Filters}.
Although this approach is based on supervised classifiers, it could also be interpreted as an anomaly detection method, see Section \ref{lbl_Discussion_AP1}.

We will rely on continuous data collection to improve our approaches and evaluate our anomaly detection method based on image rotations.
With this perspective, future work will focus on optimizing our false positive rates and unsupervised methods.

\section*{\uppercase{Acknowledgements}}

We would like to thank our industry partner Gerresheimer AG  for their cooperation and insight.

\bibliographystyle{apalike}
{\small
	\bibliography{example}}

\end{document}